\documentclass{MYsig-alternate}

\usepackage{amssymb}
\usepackage{amsmath}
\usepackage{graphicx}
\usepackage[utf8]{inputenc}
\usepackage{color}
\usepackage{algorithm}
\usepackage[noend]{algorithmic}

\usepackage{wasysym} 
\usepackage{xstring}
\usepackage{graphicx}
\usepackage{float}
\usepackage{rotating}
\usepackage{lscape}
\usepackage{colortbl}
\usepackage[table]{xcolor}
\usepackage{url}

\usepackage{amssymb,amsmath}
\usepackage{graphicx}
\usepackage{xcolor}
\usepackage{float}
\usepackage{rotating}

\newcommand{\fopt}{\ensuremath{f_\mathrm{opt}}}
\newcommand{\ftarget}{\ensuremath{f_\mathrm{t}}}

\newcommand{\bbobdatapath}{ppdatamany/} 

\graphicspath{{\bbobdatapath}}

\newcommand{\ALGOactive}{IPOP-aACM-ES}

\newcommand{\IPOPactive}{IPOP-aCMA-ES}

\def\PrevName{ACM-ES}
\def\CMA{{CMA-ES}}

\def\SAM{$^{s\ast}$}
\def\ALGOname{\SAM\hspace{-0.50ex}\PrevName}
\def\IPOPsaACMES{IPOP-\SAM\hspace{-0.50ex}aACM-ES}

\def\lifelength{lifelength}
\def\Fs{\mbox{$\widehat{f}$}}
\def\nl{\mbox{$\widehat{n}$}}
\def\nlm{\mbox{$\widehat{n_{max}}$}}
\def\CB{\mbox{$C_{base}$}}
\def\CP{\mbox{$C_{pow}$}}
\def\Cs{\mbox{$c_{sigma}$}}
\def\Xs{\mbox{$\sigma_x$}}
\def\N{\mbox{$N_{training}$}}
\def\RSVM{Ranking SVM}

\usepackage{url}

\conferenceinfo{GECCO'12,} {July 7--11, 2012, Philadelphia, Pennsylvania, USA.} 
\CopyrightYear{2012} 
\crdata{978-1-4503-1177-9/12/07} 
\begin{document}

\title{Self-Adaptive Surrogate-Assisted Covariance Matrix Adaptation Evolution Strategy}

 \numberofauthors{3}
 \author{
 \alignauthor
 Ilya Loshchilov \\
 \affaddr{TAO, INRIA Saclay}\\
 \and 
 \alignauthor
  Marc Schoenauer\\
 \affaddr{TAO, INRIA Saclay}\\ \smallskip
 \affaddr{\parbox{7cm}{LRI, Univ. Paris-Sud, Orsay, France}}\\
 \email{firstname.lastname@inria.fr}
 \and 
 \alignauthor
  Mich\`ele Sebag\\
 \affaddr{CNRS, LRI UMR 8623}\\
}
\maketitle
\begin{abstract}
This paper presents a novel mechanism to adapt surrogate-assisted population-based algorithms. 
This mechanism is applied to \PrevName, a recently proposed surrogate-assisted variant of \CMA. 
The resulting algorithm, \ALGOname, adjusts online the lifelength of the current surrogate model (the number of \CMA\ generations before learning a new surrogate) and the surrogate hyper-parameters. 

Both heuristics significantly improve the quality of the surrogate model, yielding a 
significant speed-up of \ALGOname\ compared to the \PrevName\ and \CMA\ baselines. 
The empirical validation of \ALGOname\ on the BBOB-2012 noiseless testbed demonstrates the efficiency and the scalability w.r.t the problem dimension and the population size of the proposed approach, 
that reaches new best results on some of the benchmark problems.
\end{abstract}

\category{I.2.8}{Computing Methodologies}{Artificial Intelligence}{ Problem Solving, Control Methods, and Search}

\terms{Algorithms}

\keywords{Evolution Strategies,
CMA-ES,
self-adaptation,
surrogate-assisted black-box optimization,
surrogate models,
ranking support vector machine}

\section{Introduction}
Evolutionary Algorithms (EAs) have become popular tools for optimization mostly thanks to their population-based properties and the ability to progress towards an optimum using problem-specific variation operators. A search directed by a population of candidate solutions is quite robust with respect to a moderate noise and multi-modality of the optimized function, in contrast to some classical optimization methods such as quasi-Newton methods (e.g. BFGS method \cite{BFGS1970}). Furthermore, 
many bio-inspired algorithms such as EAs, Differential Evolution (DE) and Particle Swarm Optimization (PSO) with rank-based selection are compa\-ri\-son-based algorithms, which makes their behavior invariant 
and robust under any monotonous transformation of the objective function.
Another source of robustness is the invariance under orthogonal transformations of the search space, first introduced into the realm of continuous evolutionary optimization by Covariance Matrix Adaptation Evolution Strategy (\CMA) \cite{HansenECJ01}. \CMA, winner of the Congress on Evolutionary Computation (CEC) 2005 \cite{CEC2005Garcia2009} and the Black-Box Optimization Benchmarking (BBOB) 2009 \cite{Hansen2010BBOB} competitions of continuous optimizers, has also demonstrated its robustness 
on real-world problems through more than one hundred applications \cite{CMA-ESApplications}.

When dealing with expensive optimization objectives, the well-known surrogate-assisted approaches proceed by learning a surrogate model of the objective, and using this surrogate to reduce the number of computations of the objective function in various ways. The best studied approach relies on the use of 
computationally cheap polynomial regression for the line search in gradient-based search methods, such as in the BFGS method \cite{BFGS1970}. More recent approaches rely on Machine Learning algorithms, modelling the objective function through e.g. Radial Basis Functions (RBF), Polynomial Regression, Support Vector Regression (SVR), Artificial Neural Network (ANN) and Gaussian Process (GP) a.k.a. Kriging. As could have been expected, there is no such thing as a best surrogate learning approach \cite{Jin00comparativestudies,Lim2007}. Experimental comparisons also suffer from the fact that the results depend on the tuning of the surrogate hyper-parameters.
Several approaches aimed at the adaptive selection of surrogate models during the search have been proposed \cite{Liang2008,2008:Tenne,Toscano2011}; these approaches focus on measuring the quality of the surrogate models and using the best one for the next evolutionary generation.

This paper, aimed at robust surrogate-assisted optimization, presents a surrogate-adaptation mechanism (\SAM) which can be used on top of any surrogate optimization approach. \SAM\ adapts online the 
number of generations after which the surrogate is re-trained, referred to as the surrogate \lifelength;
further, it adaptively optimizes the surrogate hyper-pa\-ra\-me\-ters using an embedded \CMA\ module. A proof of principle of the approach is given by implementing \SAM\ on top of \PrevName, a surrogate-assisted variant of \CMA, yielding the \ALGOname\ algorithm. To our best knowledge, the self-adaptation of the surrogate model within \CMA\ and by \CMA\ is a new contribution. The merits of the approach are shown as \ALGOname\ show significant improvements compared to \CMA\ and \PrevName\ on the BBOB-2012 noiseless testbed.

The paper is organized as follows.
Section \ref{state} reviews some surrogate-assisted variants of Evolution Strategies (ESs) and 
\CMA. For the sake of self-containedness, the \PrevName\ combining \CMA\ with the use of a 
Rank-based Support Vector Machine is briefly described. Section \ref{analysis} discusses the 
merits and weaknesses of \PrevName\ and suggests that the online adjustment of the surrogate hyper-parameters is required to reach some robustness with respect to the optimization objective. 
The \ALGOname\ algorithm, including the online adjustment of the surrogate \lifelength\ and 
hyper-parameters on top of \PrevName, is described in section \ref{algosection}. 
The experimental validation of \ALGOname\ is reported and discussed in section \ref{expe}. Section \ref{conclusion} concludes the paper.

\section{Surrogate Models}
\label{state}
This section discusses the various techniques used to learn surrogate models, their use within 
EAs and specifically \CMA, and the properties of surrogate models in terms of invariance 
w.r.t. monotonous transformations of the objective function \cite{runarssonPPSN06}, and orthogonal transformations of the instance space \cite{rankSurrogatePPSN10}. 

\subsection{Surrogate-assisted Evolution Strategies}
\label{sectionsurrogates}
As already mentioned, many surrogate modelling approa\-ches have been used within 
ESs and \CMA: 
RBF network \cite{Hoffmann2006IEEE}, GP \cite{Ulmer2003CEC,Emmerich06}, ANN \cite{YJin2005}, SVR \cite{Ulmer2004CEC,KramerInformatica2010}, 
Local-Weighted Regression (LWR) \cite{kernHansenMetaPPSN06,augerEvoNum2010},
\RSVM\ \cite{runarssonPPSN06,rankSurrogatePPSN10,Runarsson2011ISDA}.
In most cases, the surrogate model is used as a filter (to select $\lambda_{Pre}$ promising pre-children) and/or to estimate the objective function of some individuals in the current population. The impact of the 
surrogate, controlled by $\lambda_{Pre}$, should clearly depend on the surrogate accuracy; how to measure it ? As shown by \cite{YJin2003},
the standard Mean Square Error (MSE) used to measure a model accuracy in a regression context is 
ill-suited to surrogate-assisted optimization, as it is poorly correlated with the ability to select correct individuals. Another accuracy indicator, based on the (weighted) sum of ranks of the selected 
individuals, was proposed by \cite{YJin2003}, and used by  \cite{Ulmer2004CEC,Hoffmann2006IEEE}.

\subsection{Comparison-based Surrogate Models}
Taking advantage of the fact that some EAs, and particularly \CMA, are comparison-based algorithms, 
which only require the offspring to be correctly ranked, it thus comes naturally to learn a 
comparison-based surrogate. Compari\-son-based surrogate models, first introduced by Runarsson \cite{runarssonPPSN06}, rely on learning-to-rank algorithms, such as \RSVM\
 \cite{Joachims05}. 
Let us recall \RSVM, 
assuming the reader's familiarity with Support Vector Machines \cite{SVMbookSTC:2004}.

Let $(x_1, \ldots, x_N)$ denote an $N$-sample in instance space $X$, assuming 
with no loss of generality that point $x_i$ has rank $i$. Rank-based SVM learning \cite{Joachims05}
aims at a real-valued function $\widehat{f}$ on $X$ such that ${\widehat{f}}(x_i) < {\widehat{f}}(x_j)$ 
iff $i<j$. In the SVM framework, this goal is formalized through minimizing the 
norm of $\widehat{f}$ (regularization term) subject to the $N(N-1)/2$ ordering constraints. A more tractable formulation, also used in \cite{runarssonPPSN06,rankSurrogatePPSN10},
only involves  the $N-1$ constraints related to consecutive points, ${\widehat{f}}(x_i) < {\widehat{f}}(x_{i+1})$ for $i =1\ldots N-1$.

Using the kernel trick\footnote{
The so-called kernel trick supports the 
extension of the SVM approach from linear to non-linear model spaces, by
mapping instance space $X$ onto some {\em feature 
space} $\Phi(X)$. The actual mapping cost is avoided as the scalar product in feature space $\Phi(X)$
is computed on instance space $X$ through a {\em kernel} function $K$: $\langle \; \Phi(x),\Phi(x')\;
\rangle =_{def} K(x,x')$.}, 
ranking function $\widehat{f}$ is defined as a linear function $w$ w.r.t. 
some feature space $\Phi(X)$, i.e. ${\widehat{f}}(x) = \langle \; w, \Phi(x) \; \rangle$. 
With same notations as in \cite{SVMbookSTC:2004}, the primal minimization problem 
is defined as follows:
\begin{equation}
\label{primal}
\begin{array}{l}
\mathop{\mbox{Minimize}_{\{w, \; \xi\}} \frac{1}{2} ||w||^2 + \sum_{i=1}^N C_i\xi_i} \smallskip \\
\mbox{subj. to} \left\{
\begin{array}{l}
\langle \; w, \Phi(x_i) - \Phi(x_{i+1}) \; \rangle \ge 1 - \xi_i  \;\; (i=1,~ ...N-1)\\ 
\xi_i  \ge   0   \;\; (i=1\ldots N -1) 
\end{array}
\right.
\end{array}
\end{equation}
 where slack variable $\xi_i$ (respectively constant $C_i$) accounts 
for the violation of the $i$-th constraint (resp. the violation cost).
The corresponding dual problem, quadratic in the Lagrangian multipliers $\alpha$, 
can be solved easily by any quadratic programming solver. The rank-based surrogate $\widehat{f}$ is given as

\begin{center}
${\widehat{f}}(x)=\sum_{i=1}^{N-1} \alpha_i(K(x_i,x) - K(x_{i+1},x)) $
\end{center}

By construction, ${\widehat{f}}(x)$ is invariant to monotonous transformations of the objective function, which preserve the ranking of the training points.

\subsection{Invariance w.r.t. Orthogonal Transformations}\label{sec:prev}
As already mentioned, \CMA\ is invariant w.r.t. orthogonal transformations of the search space, 
through adapting a covariance matrix during the search. This invariance property was borrowed by 
\PrevName\ \cite{rankSurrogatePPSN10}, using the covariance matrix $C$ learned by \CMA\ within
a  Radial Basis Function (RBF) kernel $K_C$, where $C^{-1}$ is used to compute Mahalanobis distance:
\begin{equation}
\label{adaptiveKernel}
K_C(x_i,x_j) = e^{-\frac{(x_i-x_j)^t C^{-1}(x_i-x_j)}{2\sigma^2}} 
\end{equation}
For the sake of numerical stability, every training point $x$ is mapped onto $x'$ such as 
\begin{eqnarray}
\label{eq:translation}
x' = C^{-1/2}(x - m),
\end{eqnarray}
where $m$ is the mean of the current \CMA\ distribution. The standard RBF kernel with Euclidean distance is used on top of this mapping. By construction, \PrevName\ inherits from \CMA\ the property of invariance under orthogonal transformations of the search space; the use of 
the covariance matrix $C$ brings significant improvements compared to standard Gaussian kernels
after \cite{rankSurrogatePPSN10}.

\section{Discussion}\label{analysis}
\begin{figure*}[t]
\centerline{ 
	\includegraphics[width=2.6truein,height=1.8truein]{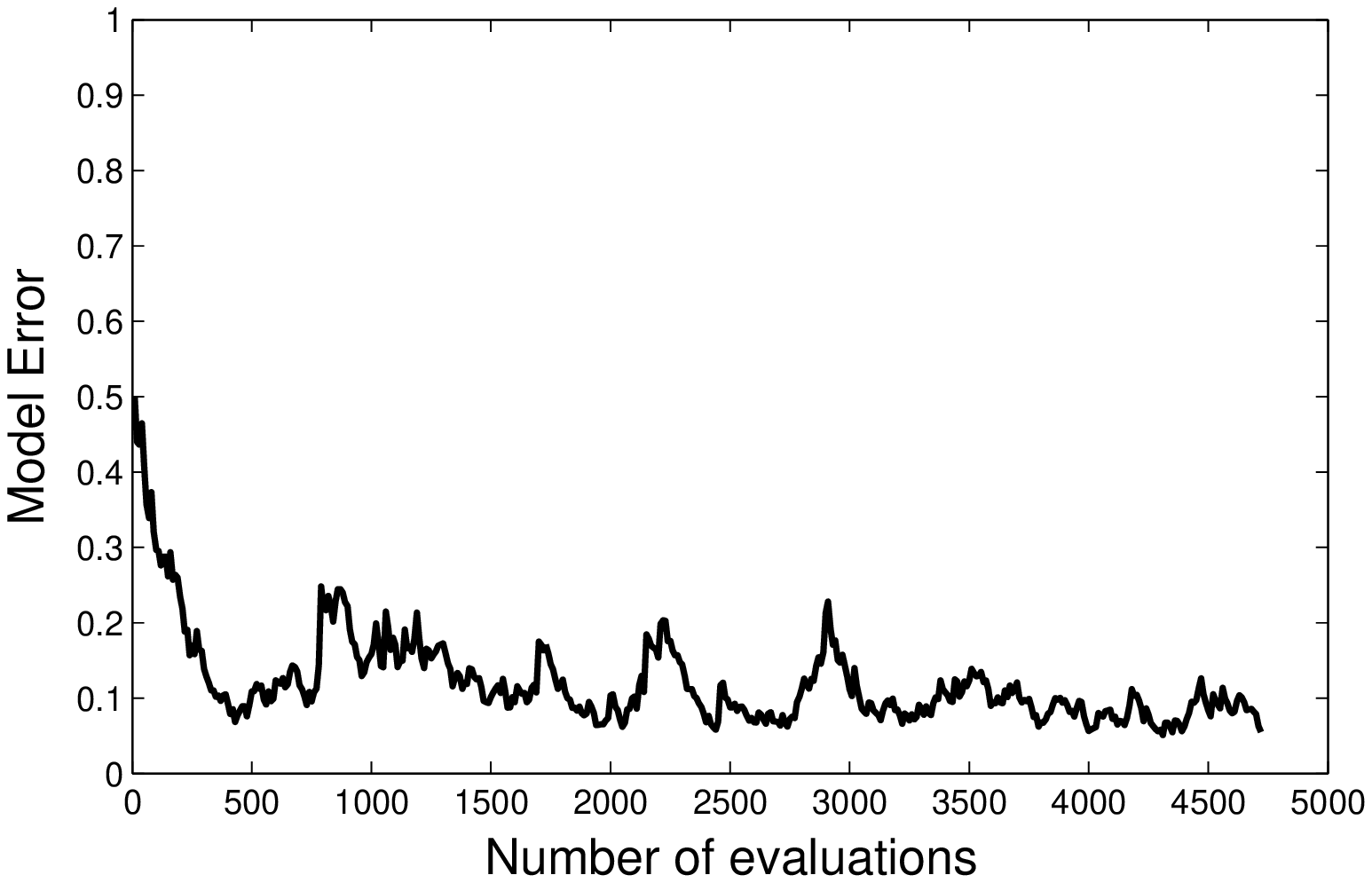} \;\;\;\;\;\;\;\;\;\;\;\;\;\;\;\;\;\;\;
	\includegraphics[height=1.8truein]{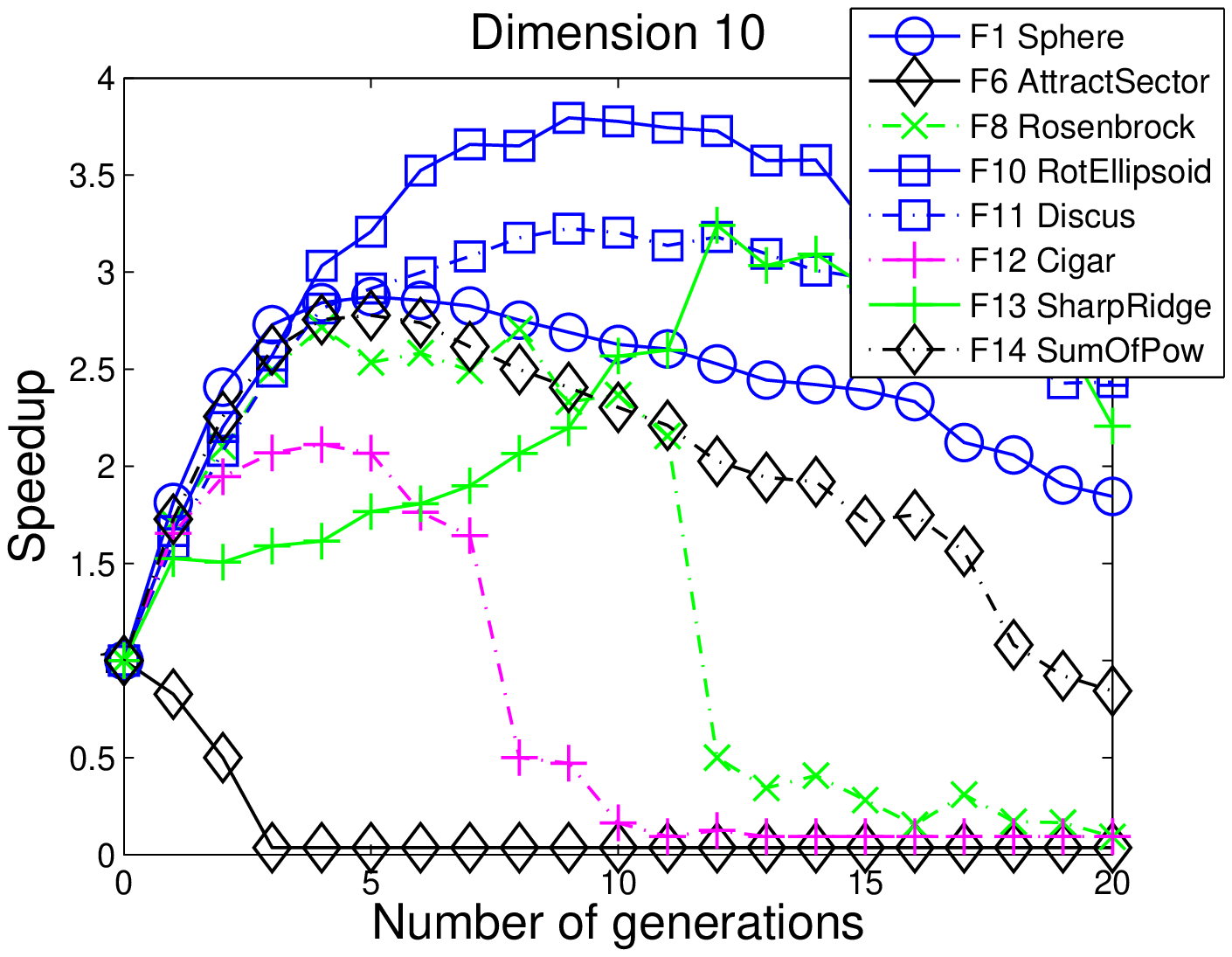} 
}
\caption{\label{fig:modelAndISTEP} Left: Rank-based surrogate error vs number of evaluations, during a representative run of active \CMA\ on 10-D Rotated Ellipsoid.
Right: 
The speedup of IPOP-aACM-ES over IPOP-aCMA-ES, where speedup = 2.0 means that IPOP-aACM-ES with a given \lifelength\ \nl\ of the  surrogate model, requires 2.0 times less computational effort SP1 (i.e. average number of function evaluations of successful runs divided by proportion of successful runs) than IPOP-aCMA-ES to reach the target objective value of $\ftarget=\fopt+10^{-8}$. 
}
\end{figure*}

This section analyzes the weaknesses of \PrevName.  
Following the characterization proposed in \cite{jinSurvey05}, \PrevName\ is a surrogate-assisted optimizer with an individual-based evolution control.
As in other pre-selection methods, at each generation \PrevName\ generates $\lambda_{Pre}$ individuals, where $\lambda_{Pre}$ is much larger than population size $\lambda$.
Then $\lambda_{Pre}$ pre-children are evaluated and ranked using surrogate model \Fs. 
The most promising $\lambda'$ pre-children are selected and evaluated using the true expensive function, yielding new points $(x,f(x))$. 
When the objective function of $\lambda'$ individuals is known, the ranking of other $\lambda-\lambda'$ points can be approximated.

While our experimental results confirm the improvements brought by \PrevName\ on some functions
(about 2-4 times faster than \CMA\ on Rosenbrock, Ellipsoid, Schwefel, Noisy Sphere and Ackley functions up to dimension 20 \cite{rankSurrogatePPSN10}), they also show a loss of performance on the 
multi-modal Rastrigin function. Complementary experiments suggest that:
\begin{enumerate}
	\item on highly multi-modal functions the surrogate model happens to suffer from a loss of accuracy; in such cases some control is required to prevent the surrogate model from misleading the search; 
	\item surrogate-assisted algorithms may require a larger population size for multi-modal problems.
\end{enumerate}

The lack of surrogate control appears to be an important drawback in \PrevName. This control should naturally reflect the current surrogate accuracy. A standard measure of the rank-based surrogate error is given as the fraction of violated ranking constraints on the test set \cite{Joachims05}. Accuracy 0 (respectively .5) corresponds to a perfect surrogate (resp. random guessing).


However, before optimization one should be sure that the model gives a reasonable prediction of the optimized function. 
Fig. \ref{fig:modelAndISTEP} (Left) illustrates the surrogate model error during a representative run of active \CMA\ (with re-training at each iteration, but without any exploitation of the model) 
on 10 dimensional Rotated Ellipsoid function.
After the first generations, the surrogate error decreases to approximately 10\%.
This better than random prediction can be viewed as a source of information about the function which can be used to improve the search.

Let \nl\ denote the number of generations a surrogate model is used, referred to as surrogate \lifelength. In so-called generation-based evolution control methods \cite{jinSurvey05}, the surrogate \Fs\ is directly optimized for \nl\ generations, without requiring any expensive objective computations. The following generation considers the objective function $f$, and yields instances to enrich the training set, relearn or refresh the surrogate and adjust some parameters of the algorithm. 
The surrogate lifelength \nl\ is fixed or adapted.

The impact of \nl\
is displayed on Fig. \ref{fig:modelAndISTEP} (Right), showing the speedup reached by direct surrogate
optimization on several 10 dimensional benchmark problems {\em vs} the number of generations \nl\ the surrogate is used. A factor of speedup 1.7 is obtained for \nl$=1$ on the Rotated Ellipsoid function, close to the 
optimal speedup 2.0. A speedup ranging from 2 to 4  is obtained for IPOP-aCMA-ES with surrogates for $\nl ~in ~[5, 15]$. 
As could have been expected again, the optimal value of \nl\ is problem-dependent and widely varies. In the case of the Attractive Sector problem for instance, the surrogate model is not useful and $\nl=0$ should be used (thus falling back to the original aCMA-ES with no surrogate) to prevent the surrogate from misleading the search.




\def\thopt{\mbox{$\theta_{opt}$}}
\def\thsur{\mbox{$\theta_{sur}$}}
\def\A{\mbox{$\cal A$}}
\section{Self-adaptive surrogate-based CMA-ES}\label{algosection}

In this section we propose a novel surrogate adaptation mechanism which can be used in principle 
on top of any iterative population-based optimizer without requiring any significant modifications thereof.
The approach is illustrated on top of \CMA\ and \PrevName. The resulting algorithm, \ALGOname, maintains a global hyper-parameter vector $\theta = (\thopt,\thsur, \nl, \A, \alpha)$, where:\\
$\bullet$ \thopt\ stands for the optimization parameters of the CMA-ES used for expensive function optimization; \\
$\bullet$ \thsur\ stands for the optimization parameters of the CMA-ES used for surrogate model hyper-parameters optimization; \\
$\bullet$ \nl\ is the number of optimization generations during which the current surrogate model is used;\\
$\bullet$ \A\ is the archive of all points $(x_i,f(x_i))$ for which the true objective function has been
computed, exploited to train the surrogate function; \\
$\bullet$ $\alpha$ stands for the surrogate hyper-parameters. \\
All hyper-parameters are indexed by the current generation $g$;  by abuse of notations, the subscript $g$ is omitted when clear from the context.

The main two contributions of the paper regard the adjustment of the surrogate hyper-parameters (section \ref{sec:th}) and 
of the surrogate \lifelength\ \nl\ (section \ref{sec:nl}).

\def\gs{\mbox{$g_{start}$}}
\def\G{\textit{GenCMA}}
\def\thh{\mbox{$\theta_{h}$}}
\subsection{Overview of \ALGOname}
Let \G($h$,\thh,\A) denote the elementary optimization module (here one generation of \CMA)
where $h$ is the function to be optimized (the true objective $f$ or the surrogate \Fs), $\thh$ denotes the current optimization parameters (e.g. \CMA\ step-size and covariance matrix) associated to $h$, and \A\ is the archive of $f$.
After each call of \G, optimization parameters \thh\ are updated; and if \G\ was called with the true
objective function $f$, archive \A\ is updated and augmented with the new points $(x,f(x))$.
Note that \G\ can be replaced by any black-box optimization procedure, able to update its own optimization parameters and the archive.

\ALGOname\ starts by calling \G\ for \gs\ number of generations with the true objective $f$, where \thopt\ and \A\ are respectively initialized to the default parameter of \CMA\ and the empty set
(lines \ref{algoinitonfbegin}-\ref{algoinitonfend}). 
In this starting phase, optimization parameter \thopt\ and archive \A\ are updated in each generation.

\begin{figure}[t]
\centerline{ 
	\includegraphics[width=3.4truein]{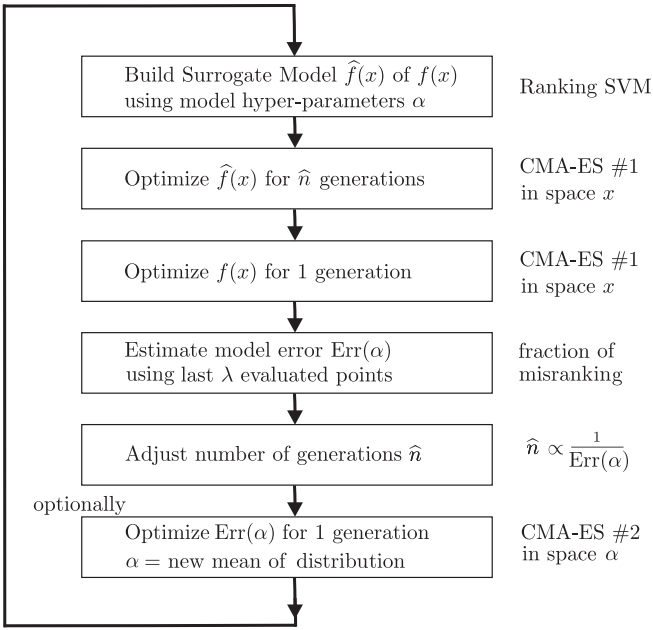}
}
\caption{ Optimization loop of the \ALGOname. 
}
\label{figscheme}
\end{figure}

Then \ALGOname\ iterates a three-step process (Algorithm \ref{algoalgo}, illustrated on Fig. \ref{figscheme}):
\begin{itemize}
 \item[1] learning surrogate \Fs\ (procedure BuildSurrogateModel, line \ref{algobuildmodel}; section \ref{sec:th});
\item[2] Optimizing surrogate \Fs\ during \nl\ generations (lines \ref{algoCycle1}-\ref{algoCycle2}). This step classically calls \G(\Fs,\thopt,\A) for \nl\ consecutive generations; \thopt\ is updated accordingly while \A\ is unchanged since this step does not involve any computation of the expensive $f$.
\item[3] Adjusting the surrogate \lifelength~ \nl\ (section \ref{sec:nl}). 
\end{itemize}

\def\tht{\mbox{$\alpha$}}
\def\Fst{\mbox{\Fs$_\alpha$}}
\def\err{{\mbox{Err}}}
\def\Fsg{\Fs$_g$}
\def\errth{\mbox{$\tau_{err}$}}

\begin{algorithm}[tb!]
\caption{\ALGOname}
\label{algoalgo}
\begin{algorithmic}[1]
\STATE{$ g \leftarrow 0;$ $\err \leftarrow 0.5;$ $\A_{g} \leftarrow \emptyset; $} 
\STATE{ $\thopt$ $\leftarrow$ InitializationCMA(); }	\COMMENT{	to optimize $f(x)$ }	\label{algoinit}
\STATE{ $\thsur$ $\leftarrow$ InitializationCMA(); }	\COMMENT{	to optimize $h(\alpha)$ } \label{algoinitH}
\REPEAT		\label{algoinitonfbegin}
		\STATE{ $\left\{ \thopt, \A_{g+1} \right\}$ $\leftarrow$ GenCMA($f$,\thopt, $\A_{g}$); }
		\STATE{ $ g \leftarrow g + 1;$ }
\UNTIL{$g = g_{start}$ ;}		\label{algoinitonfend}
\REPEAT 
	\STATE{ $\widehat{f}(x)$ $\leftarrow$ BuildSurrogateModel($\alpha$, $\A_{g}$, $\thopt$); }		\label{algobuildmodel}
	\STATE{ $g_{prev} \leftarrow g$;}
	\FOR{$i = 1,\ldots,\widehat{n}$}	\label{algoCycle1}
			\STATE{ $\left\{ \thopt, \A_{g+1}=\A_{g} \right\}$ $\leftarrow$ GenCMA($\widehat{f}$,$\thopt$, $\A_{g}$); }
			\STATE{ $ g \leftarrow g + 1;$ }				\label{algoCycle2}
	\ENDFOR	
  \STATE{ $\left\{ \thopt, \A_{g+1} \right\}$ $\leftarrow$ GenCMA($f$,$\thopt$, $\A_{g}$); }		\label{algoonf}
	\STATE{ $ g \leftarrow g + 1;$ }
	\STATE{ $\err (\alpha)$ $\leftarrow$ MeasureSurrogateError($\widehat{f}$,$\thopt$); }		\label{algomodelerror}
	\STATE{ $\err \leftarrow (1-\beta_\err)\err + \beta_\err \err(\alpha) $; }		\label{algoerrorsmoothing}
	\STATE{ $\widehat{n} \leftarrow \left\lfloor \frac{\errth - \err}{\errth} \nlm \right\rfloor;$ }	\label{algongen}
	\STATE{ // adjust surrogate hyperparameters }
			\STATE{ $\thsur$ $\leftarrow$ GenCMA($\err$,$\thsur$); }	\label{algogenH}
			\STATE{ $ \alpha  \leftarrow \thsur.m$;}	\label{algoupdateH}
\UNTIL{stopping criterion is met ;}
\end{algorithmic}
\end{algorithm}

\begin{algorithm}[tb!]
\label{alfunch}
\caption{Objective function $\err(\alpha)$ of surrogate model}
\begin{algorithmic}[1]
\STATE{ Input: $\alpha$ }
\STATE{ $\widehat{f}(x)$ $\leftarrow$ BuildSurrogateModel($\alpha$, $\A_{g_{prev}}$, $\theta_{sur,g_{prev}}$); }
\STATE{ $\err (\alpha)$ $\leftarrow$ MeasureSurrogateError($\widehat{f}$, $\theta_{opt,g_{prev}}$); }
\STATE{ Output: $\err (\alpha)$; }
\end{algorithmic}
\end{algorithm}

\subsection{Learning a Surrogate and Adjusting its Hy\-per-parameters}\label{sec:th}
The surrogate model learning phase proceeds as in \PrevName\ (section \ref{sec:prev}). 
\G($f$,\thopt,\A) is launched for one generation with the true objective $f$, updating and augmenting archive \A\ with new $(x,f(x))$ points. 

\Fs\ is built using \RSVM\ \cite{Joachims05} with archive \A\ as training set, where the SVM kernel is tailored using the current optimization parameters \thopt\ such as covariance matrix $C$.

The contribution regards the adjustment of the surrogate hyper-parameters $\alpha$ (e.g. the number and selection of the training points in \A; the weights of the constraint violations in \RSVM, section \ref{sec:prev}),  which are adjusted to optimize the quality of the surrogate \err\ (Eq. \ref{modelerror}). Formally,
to each surrogate hyper-parameter vector \tht\ is associated a surrogate error \err(\tht) defined as follows: hyper-parameter \tht\ is used to learn surrogate \Fst\ using \A$_{g-1}$ as training set, and \err(\tht) is set to the ranking error of \Fst, using the most recent points (\A$_{g} - \A_{g-1}$) 
as test set.

Letting $\Lambda$ denote the test set and assuming with no loss of generality that the points in $\Lambda$ are ordered after $f$, \err(\Fst) is measured as follows (procedure MeasureSurrogateError, line \ref{algomodelerror}): 

\begin{equation}
\label{modelerror}
\err(\tht) = \frac{2}{|\Lambda| (|\Lambda|-1)}\sum_{i=1}^{|\Lambda|}\sum_{j=i+1}^{|\Lambda|} w_{ij} . 1_{\Fst, i, j}
\end{equation}
 where $1_{\Fst, i,j}$ holds true iff \Fst\ violates the ordering constraint on pair (i,j).
In all generality, the surrogate error can be tuned using weight coefficients $w_{ij}$ to reflect the
relative importance of ordering constraints. Only $w_{ij}=1$ will be used in the remainder of the paper. For a better numerical stability, the surrogate error is updated using additive relaxation, with relaxation constant $\beta_\err$ (lines \ref{algomodelerror}-\ref{algoerrorsmoothing}).

Finally, \hfill the \hfill elementary \hfill optimization \hfill module\\ \G(\err,\thsur) (in this study we do not use archive parameter here) is launched for one generation (line \ref{algogenH}), and the 
 mean of the \CMA\ mutation distribution is used (line \ref{algoupdateH}) as surrogate hyper-parameter vector in the next surrogate building phase 
(line \ref{algobuildmodel}). 

\subsection{Adjusting Surrogate Lifelength}
\label{sec:nl}
\def\nlg{\mbox{\nl$_g$}}
\def\Fsg{\Fs$_{g-1}$}
\def\errth{\mbox{$\tau_{err}$}}

\begin{figure}[t]
\centerline{ 
	\includegraphics[width=2.5truein]{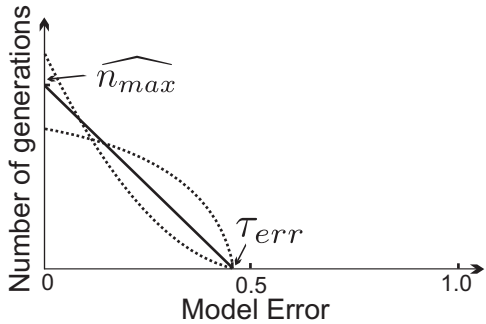}
}
\caption{Number of generations \nl\ versus surrogate error \err. Linear interpolation
(bold curve) has been used in the experimental validation.
}
\label{figmodelerror}
\end{figure}

Lifelength \nlg\ is likewise adjusted depending on the error made by the previous surrogate \Fsg\ on the new archive points ($\A_{g} - \A_{g-1}$).
If this error is 0, then \Fsg\ is perfectly accurate and could have been used for some more generations before learning \Fs$_{g}$. In this case  lifelength \nlg\ is set to the maximum value \nlm, which
corresponds to the maximum theoretical speedup of the \ALGOname.\\
If the error is circa .5, 
surrogate \Fsg\ provides no better indications than random guessing and thus misleads the optimization; \nlg\ is set to 0.
More generally, considering an error threshold \errth, \nl\ is adjusted between \nlm\
and 0, proportionally\footnote{Complementary experiments omitted for brevity, show that the best adjustment of \nl\ depending on the surrogate error is again problem-dependent.} to the ratio between the actual error and the error threshold \errth\ (line \ref{algongen}, bold curve on Fig. \ref{figmodelerror}).

\section{Experimental validation}\label{expe}
The experimental validation of the approach proceeds by comparing the 
performance of \ALGOname\ to the original \cite{HansenECJ01} and active \cite{2006:JastrebskiArnold} \CMA\ versions, considering the restart scenario with 
increasing population size (IPOP \cite{AugerIPOPCMA2005,HansenActiveIPOPCMA2010}).

The active IPOP-aCMA-ES \cite{HansenActiveIPOPCMA2010} with weighted negative covariance matrix update is found to perform equally well
or better than IPOP-CMA-ES, which is explained as it more efficiently exploits the information of the worst $\lambda/2$ points. We use IPOP-aCMA-ES as challenging baseline, more difficult to speed up than the original IPOP-CMA-ES.

Specifically, \ALGOname\ is validated on the noiseless BBOB testbed by comparing IPOP-aACM-ES with fixed hyper-parameters, and \IPOPsaACMES\ with online adaptation of hyperparameters of the surrogate model\footnote{
For the sake of reproducibility we used the Octave/MatLab source code of IPOP-\CMA\ with default
parameters, available from its author's page, with the active flag set to 1. The  \ALGOname\ source code is available at \\ \url{https://sites.google.com/site/acmesgecco/}.}.
 
After detailing the experimental setting, this section reports on the offline tuning of the number \N\ of 
points used to learn the surrogate model, and the online tuning of the surrogate hyper-parameters.

\subsection{Experimental Setting}
The default BBOB stopping criterion is reaching target function value $\ftarget=\fopt+10^{-8}$. 
\RSVM\ is trained using the most recent $N_{training}$ points (subsection \ref{ntrainingpoints}); 
its stopping criterion is arbitrarily set to a maximum 
number of $1000 N_{training}$ iterations of the quadratic programming solver.

After a few preliminary experiments, 
the \RSVM\ constraint violation weights (Eq. \ref{primal}) are set to 
\[C_i=10^{\CB}(\N-i)^{\CP}\]
 with $\CB=6$ and $\CP=3$ by default; the cost of constraint violation is thus cubically higher for top-ranked samples. The $\sigma$ parameter of the RBF kernel is set to $\sigma=\Cs \Xs$, where \Xs\ is the 
dispersion of the training points (their average distance after translation, Eq. \ref{eq:translation})
and \Cs\ is set to 1 by default.
The number $g_{start}$ of \CMA\ calls in the initial phase is set to 10, the maximum lifelength \nlm\ of a surrogate model is set to 20. The error threshold \errth\ is set to .45 and the error relaxation factor is set to .2.

The surrogate hyper-parameters \thsur\ are summarized in Table \ref{tab:sur}, with offline tuned value (default for IPOP-aACM-ES) and their range of variation for online tuning, (where $d$ stands for the problem dimension). Surrogate hyper-para\-me\-ters are optimized with a population size 20 (20 surrogate models), where the \err\ function associated to a hyper-parameter vector is measured on the most recent $\lambda$ points in archive \A, with $\lambda$ the current optimization population size.
\begin{table}
\centerline{\begin{tabular}{l|r|r}
  Parameter & Range for online tuning & Offline tuned value\\ \hline
\N & $\left[ 4d, 2(40 + \left\lfloor  4 d^{1.7} \right\rfloor) \right]$ & $40 + \left\lfloor  4 d^{1.7} \right\rfloor$\\
\CB & $\left[ 0, 10 \right]$ & 6\\
\CP & $\left[ 0, 6 \right]$ & 3\\
\Cs & $\left[ 0.5, 2 \right]$ & 1\\
\end{tabular}}
\caption{Surrogate hyper-parameters, default value and range of variation}
\label{tab:sur}
\end{table}

\subsection{Offline Tuning: Number of Training Points}
\label{ntrainingpoints}
It is widely acknowledged that the selection of the training set is an essential ingredient of surrogate
learning \cite{jinSurvey05}. After some alternative experiments omitted for brevity, the training set includes simply the most recent \N\ points in the archive. The study thus focuses on the tuning of \N. 
Its optimal tuning is of course problem- and surrogate learning algorithm-dependent. Several tunings
have been considered in the literature, for instance for 10-di\-men\-sio\-nal problems: 
3$\lambda$ for SVR \cite{Ulmer2004CEC}; 
30 for RBF \cite{Ulmer2003CEC};
50 for ANN \cite{YJin2005};
$\lambda,2\lambda$ for \RSVM \cite{runarssonPPSN06,Runarsson2011ISDA};
$\frac{d(d+1)}{2}+1=66$ for LWR in the lmm-CMA-ES \cite{augerEvoNum2010};
$70\sqrt{d}=221$ for \RSVM\ in the ACM-ES \cite{rankSurrogatePPSN10}.

In all above cases but \PrevName, the surrogate models aim at local approximation. These approaches might
thus be biased toward small \N\ values, as a small number of training points are required to yield good local models (e.g. in the case of the Sphere function), and small \N\ values positively contribute to the speed-up. It is suggested however that the Sphere function might be misleading, regarding the 
optimal adjustment of \N.

\begin{figure}[t]
\centerline{ 
	\includegraphics[width=2.7truein]{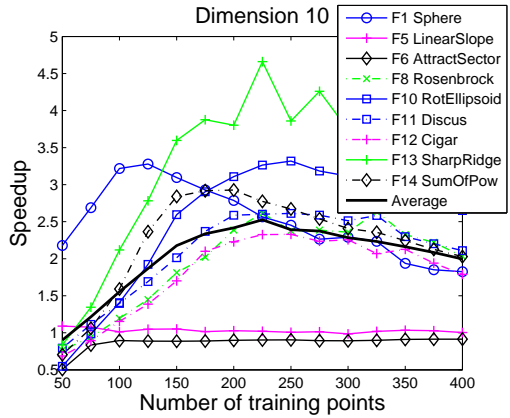}
}
\caption{The speedup of \ALGOactive\ over \IPOPactive\ w.r.t. (fixed) number of training points.
}
\label{graph1trdim}
\end{figure}

Let us consider the surrogate speed-up of IPOP-aACM-ES w.r.t. IPOP-CMA-ES depending on (fixed) \N, on uni-modal benchmark problems from the BBOB noiseless testbed (Fig. \ref{graph1trdim} for $d=10$). While the optimal speed-up varies from 2 to 4, the actual speed-up 
strongly depends on the number \N\ of training points. 

Complementary experiments on $d$-dimensional problems with $d=2,5,10,20,40$ (Fig. \ref{graph1trdim}) yield to propose an average best tuning of \N\ depending on dimension $d$:
\begin{equation}
\N =\left\lfloor 40+4d^{1.7}\right\rfloor
\label{eq:N} 
\end{equation}
Eq. (\ref{eq:N}) is found to empirically outperform the one proposed in \cite{rankSurrogatePPSN10}
($\N=\left\lfloor 70\sqrt{d}\right\rfloor$), which appears to be biased to 10-dimensional problems, and underestimates the number of training points required in higher dimensions. Experimentally however, \N\ must super-linearly increase with $d$; eq. (\ref{eq:N}) states that for large $d$ the number of training points should triple when $d$ doubles. 

Further, Fig. \ref{graph1trdim} shows that the optimal \N\ value is significantly smaller for the Sphere
function than for other functions, which experimentally supports our conjecture that the 
Sphere function might be misleading with regard to the tuning of surrogate hyper-parameters.

\begin{figure}[t]
\centerline{ 
	\includegraphics[width=3.25truein]{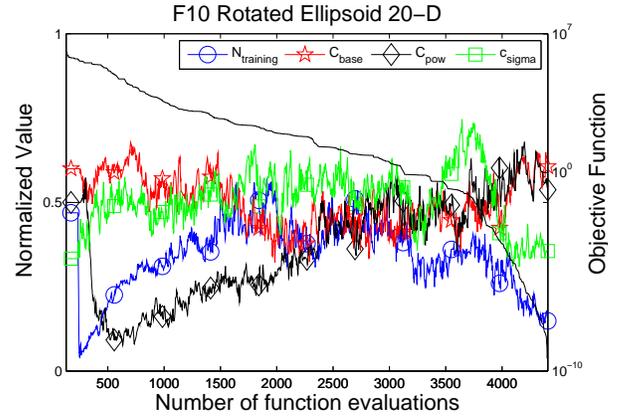}
}
\caption{The median trajectories of normalized surrogate hyper-parameters estimated on 15 runs of the \IPOPsaACMES\ on Rotated Ellipsoid 20-D.
}
\label{hyperOptimization}
\end{figure}

\subsection{Online Tuning: Surrogate Hyper-parameters}
\label{hyperOptim}
The IPOP-\ALGOname\ achieves the online adaptation of the surrogate hyper-parameters 
within a specified range (Table \ref{tab:sur}), yielding the surrogate hyper-parameter values to be used in the next surrogate learning step. 

Note that a surrogate hyper-parameter individual might be non-viable, i.e. if it does not enable 
to learn a surrogate model (\RSVM\ fails due to an ill-conditioned kernel). Such non-viable individual is heavily penalized (\err(\tht) $>>1$). In case no usable hyper-parameter 
individual is found (which might happen in the very early generations as it is shown on Fig. \ref{hyperOptimization}), 
\thsur\ is set to its default value. 

The online adaptation of surrogate hyper-parameters however 
soon reaches usable hyper-parameter values. The trajectory of the surrogate hyper-parameter values vs the number of generations is depicted in Fig. \ref{hyperOptimization}, normalized in $[0,1]$ and 
considering the median out of 15 runs optimizing 20 dimensional Rotated Ellipsoid function.

The trajectory of \N\ displays three stages. In a first stage, \N\ increases as the overall number of evaluated points (all points are required to build a good surrogate). In a second stage, \N\ reaches a plateau; its value is close to the one found by offline tuning (section \ref{ntrainingpoints}). In a third stage, \N\ steadily decreases. This last stage is explained as \CMA\ approaches the optimum of $f$ and 
gets a good estimate of the covariance matrix of the Ellipsoid function. At this point the optimization
problem is close to the Sphere function, and 
a good surrogate can be learned from comparatively few training points. 

The trajectories of other surrogate hyper-parameters are more difficult to interpret, although they
clearly show non-random patterns (e.g. \CP). 

\subsection{Comparative Performances}
\begin{figure*}

\centerline{ 
	\includegraphics[height=1.8truein]{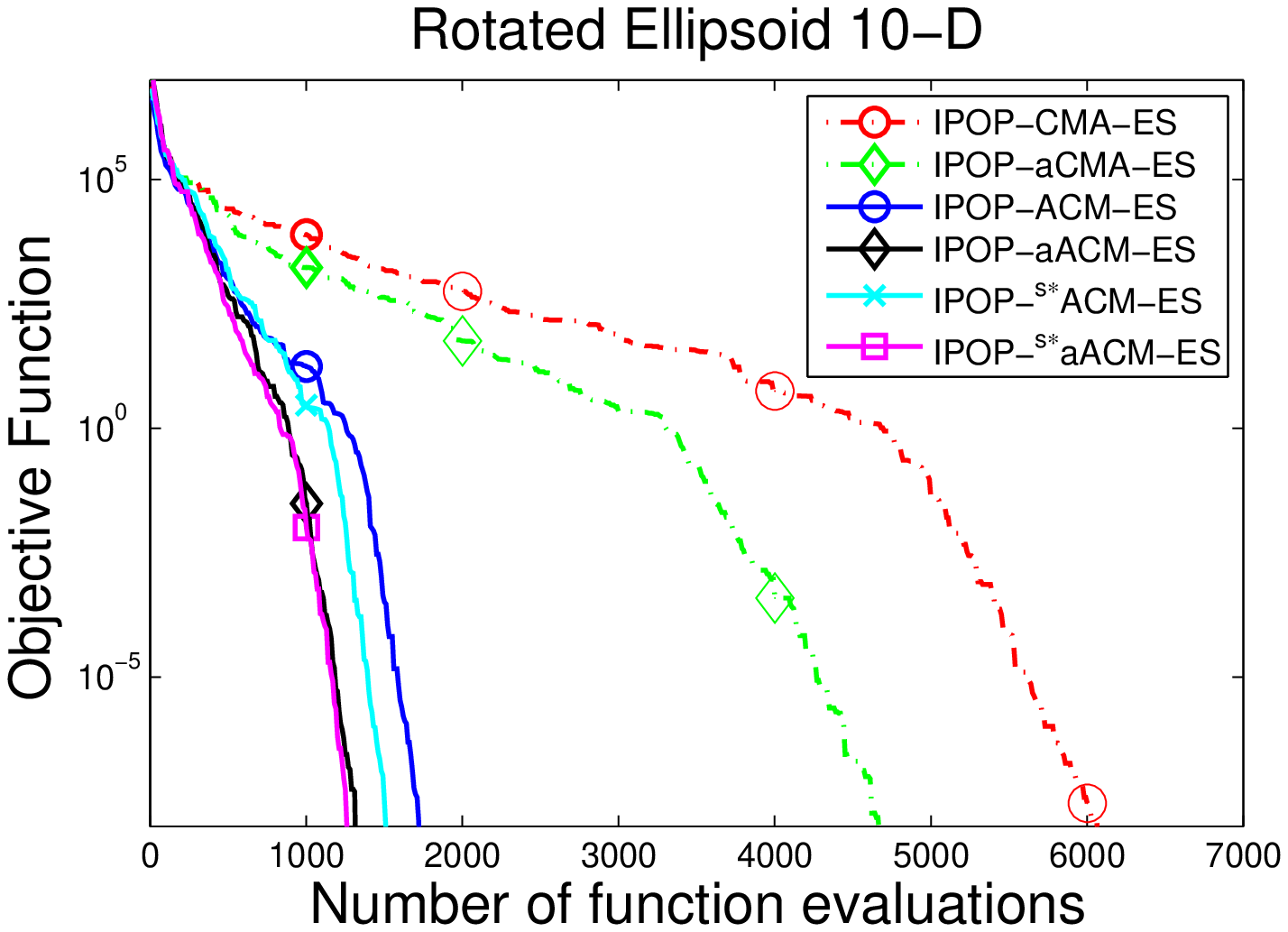} \;\;\;\;\;\;\;\;\;\;\;\;\;\;\;\;\;\;\;\;\;\;\;
  \includegraphics[height=1.8truein]{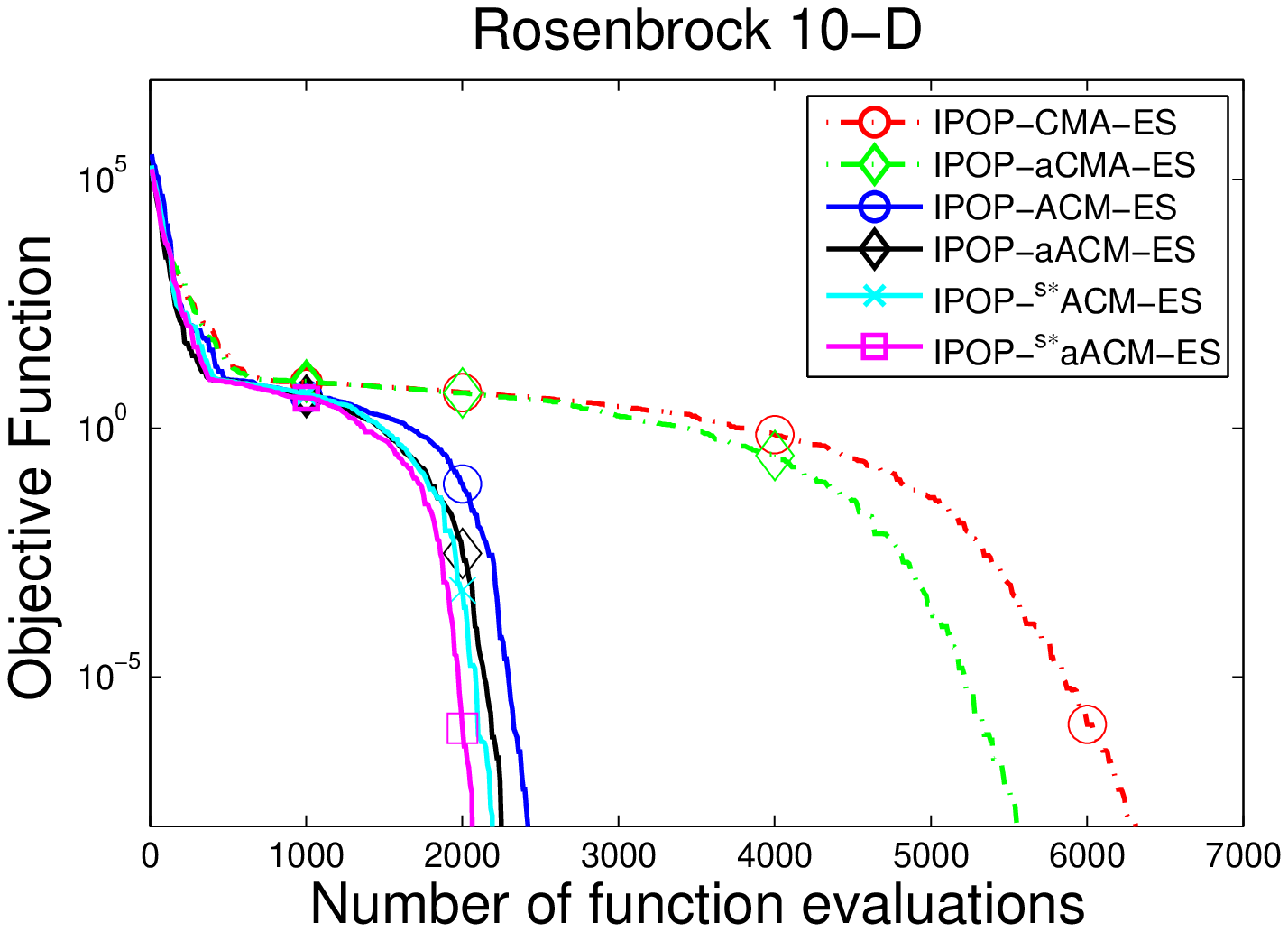}
}

\caption{Comparison of the proposed surrogate-assisted versions of the original and active IPOP-CMA-ES algorithms on 10 dimensional Rotated Ellipsoid (Left) and Rosenbrock (Right) functions. The trajectories show the median of 15 runs. 
}
\label{graph2ellirosen}
\end{figure*}

The comparative performance of \ALGOname\ combined with the original and the active variants of IPOP-\CMA\
is depicted on Fig. \ref{graph2ellirosen}, on the 10-d Rotated Ellipsoid (Left) and Rosenbrock (Right) functions. In both cases, the online adaptation of the surrogate hyper-parameters yields a quasi constant speed-up, witnessing the robustness of \ALGOname.
On the Ellipsoid function, the adaptation of the covariance matrix is much faster than for the baseline, yielding same convergence speed as for the Sphere function.
On the Rosenbrock function the adaptation is also much faster, although there is clearly room for improvements. 

The performance gain of \ALGOname, explained from the online adjustment of the surrogate hyper-parameters, in particular \N, confirms the fact that the appropriate surrogate  hyper-parameters
vary along search, and can be adjusted based on the accuracy of the current surrogate model.
Notably, IPOP-\ALGOname\ almost always outperforms IPOP-\PrevName, especially for $d>10$.


\subsection{Scalability w.r.t Population Size}
\label{largePop}
\def\L{\mbox{$\lambda_{default}$}}
The default population size $\lambda_{default}$ is suggested to be the only \CMA\ parameter to 
possibly require manual tuning. Actually, \L\ is well tuned for uni-modal problems and only depends on the problem dimension.
Increasing the population size does not decrease the overall number of function evaluations needed to reach an optimum in general. Still, it allows one to reach the optimum after fewer generations. Increasing the population size and running the objective function computations in parallel 
is a source of speed-up, which raises the question of \ALGOname\ scalability w.r.t. the population size.

\begin{figure}
\centerline{ 
	\includegraphics[height=1.9truein]{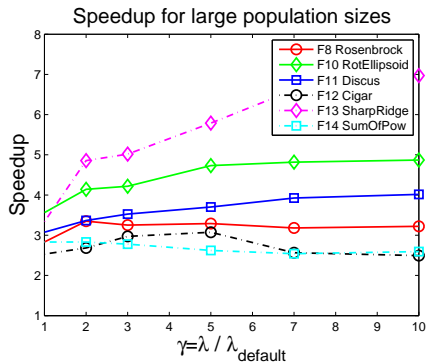}
}
\caption{Speedup of the \IPOPsaACMES\ over IPOP-aCMA-ES for large population sizes $\lambda = \gamma \lambda_{default}$ on 10-D problems. 
}
\label{graphLargePop}
\end{figure}

Fig. \ref{graphLargePop} shows the speedup of the \IPOPsaACMES\ compared to IPOP-aCMA-ES for unimodal 10 dimensional problems, when the population size $\lambda$ is set to $\gamma$ times the default population size \L.
For F8 Rosenbrock, F12 Cigar and F14 Sum of Different Powers the speedup remains almost constant and independent of $\gamma$,
while for F10 Rotated Ellipsoid, F11 Discus and F13 Sharp Ridge, it even increases with $\gamma$. We believe that with a larger population size, ``younger'' points are used to build the surrogate model, that is hence more accurate.

The experimental evidence suggests that \ALGOname\ can be applied on top of 
parallelized versions of \IPOPsaACMES, while preserving or even improving its speed-up. Note that
the same does not hold true for all surrogate-assisted methods; for instance in trust region methods, one needs to sequentially evaluate the points. 

It is thus conjectured that further improvements of \CMA\ (e.g. refined parameter tuning, noise handling)
will translate to \ALGOname, without degrading its speed-up.

\section{Conclusion and Perspectives}
\label{conclusion}

This paper presents a generic framework for adaptive sur\-rogate-assisted optimization, which can in principle be combined with any iterative population-based optimization, and surrogate learning, algorithms. This framework has been instantiated on top of surrogate-assisted \PrevName, using \CMA\ as optimization algorithm and \RSVM\ as surrogate~ learning~ algorithm.~  
The~ resulting~ algorithm, $\mbox{\ALGOname}$, inherits from \CMA\ and \PrevName\ the property of invariance w.r.t. monotonous transformations of the objective function and orthogonal transformations of the search space. 

The main contribution of the paper regards the online adjustment of i) the number \nl\ of generations a surrogate model is used, called surrogate lifelength; ii) the surrogate hyper-parameters controlling the surrogate learning phase. 
The surrogate lifelength is adapted depending on the quality of the current surrogate model; the higher the quality, the longer the next surrogate model will be used. The adjustment of the surrogate hyper-parameters is likewise handled by optimizing them w.r.t. the quality of the surrogate model, without 
requiring any prior knowledge on the optimization problem at hand.

\IPOPsaACMES\ was found to improve on IPOP-aCMA-ES with a speed-up ranging from 2 to 3 
on uni-modal $d$-dimensional functions from the BBOB-2012 noiseless testbed, with dimension $d$ ranging from 2 to 40. On multi-modal functions, \IPOPsaACMES\ is equally good or sometimes better than IPOP-aCMA-ES, although the 
speed-up is less significant than for uni-modal problems. 
Further, \IPOPsaACMES\ also improves on IPOP-aCMA-ES on problems with moderate noise from BBOB-2012 noisy testbed.
All these results as well as the computational complexity of the algorithm are discussed in details in BBOB-2012 workshop papers \cite{BBOB2012noiseless} and \cite{BBOB2012noisy}.

A long term perspective for further research is to better handle multi-modal and noisy functions.
A shorter-term perspective is to consider a more comprehensive surrogate learning phase, involving 
a portfolio of learning algorithms and using the surrogate hyper-parameter optimization phase to 
achieve portfolio selection.
Another perspective is to design a tighter coupling of the surrogate learning phase, and the 
\CMA\ optimization, e.g. using the surrogate model \Fs\ to adapt the \CMA\ hyper-parameters during
the optimization of the expensive objective $f$. 

\section{Acknowledgments}
The authors would like to acknowledge Anne Auger, Zyed Bouzarkouna, Nikolaus Hansen and Thomas P. Runarsson for their valuable discussions.
This work was partially funded by FUI of System@tic Paris-Region ICT cluster through contract DGT 117 407 {\em Complex Systems Design Lab} (CSDL). 
\bibliographystyle{abbrv}

\end{document}